\documentclass{article}
\usepackage{spconf,amsmath,graphicx,url,bbm}
\usepackage[caption=false]{subfig}
\usepackage{color}

\title{Can you find a face in a HEVC bitstream?}
\name{{Saeed Ranjbar Alvar, Hyomin Choi, and Ivan V. Baji\'c}}
\address{School of Engineering Science, Simon Fraser University, Burnaby, BC, Canada %\\
}
\begin{document}
%\ninept
%
\maketitle
\begin{abstract}
Finding faces in images is one of the most important tasks in computer vision, with applications in biometrics, surveillance, human-computer interaction, and other areas. In our earlier work, we demonstrated that it is possible to tell whether or not an image contains a face by only examining the HEVC syntax, without fully reconstructing the image. In the present work we move further in this direction by showing how to localize faces in HEVC-coded images, without full reconstruction. We also demonstrate the benefits that such approach can have in privacy-friendly face localization.   
\end{abstract}
%

% Some sentences and phrases to use; 
%the detection pipeline consists of a single forward pass through the network.

\section{Introduction}
\label{sec:intro}
Finding faces in images is one of the most important tasks in computer vision~\cite{face_detection_survey_2015}, with applications in biometrics, surveillance, human-computer interaction, and other areas. %There are two topics related to the problem of finding faces. First problem is to detect whether there is a face in the given images which we call it detection task. The other problem is to find the location of the faces in case the tested image includes a face which we refer to it as localization. In most of the papers in the literature face detection is referred to the combination of detection and localization.
Recent advances in Deep Neural Networks (DNN) have broken new ground in this field~\cite{cascade, facial_parts, hyperface, tiny}; modern approaches achieve well over 90\% true positive rate on popular benchmark datasets such as FDDB~\cite{FDDB}. However, real-world deployment of these technologies has lagged behind research advances for several reasons. One is the computational resources needed to run advanced face detection on a large scale, especially on high-resolution images. Another reason is privacy concerns. If a vision system can find a face in the image, it might also be able to recognize that face. This idea makes many people uncomfortable. In this paper, we describe a way to find faces in images that requires less computation and offers higher privacy protection than conventional approaches.

In our previous work~\cite{MIPR}, we asked if it is possible to \emph{tell a face from an HEVC bitstream}. That is to say, is it possible to distinguish images containing faces from those that do not, just from the High Efficiency Video Coding (HEVC)~\cite{HEVC}  syntax? We gave a constructive answer to that question by designing a Convolutional Neural Network (CNN)-based face detector for HEVC-coded images that performed equally well, on average, as a more conventional pixel-domain face detector that was also based on a CNN. We refer to this problem as face \emph{detection}, in line with the common use of the term ``detection'' in statistical signal processing~\cite{Kay_Detection}. The benefit of face detection directly from the bitstream is that full image reconstruction can be avoided, which saves over 60\% of HEVC decoding time, on average, across various image resolutions~\cite{MIPR}.  

In the present work, we extend this approach to face \emph{localization} by showing that it is possible not only to detect faces, but also find where they are in HEVC-coded images without full image reconstruction. We also demonstrate the potential of this approach in privacy protection. Privacy-friendly visual analytics are becoming increasingly important with the growth of public awareness of the widespread use of private data for commercial (and sometimes illegal) purposes. A recent proposal on this topic~\cite{hindering_GlobalSIP2017} advocates modifying the face region in an image in order to hinder face detection and thereby also hinder face recognition. Our approach is different. We can scramble transform coefficients over the entire image, without knowing beforehand where the faces are. Due to scrambling, conventional face detectors' performance is hindered, similar to~\cite{hindering_GlobalSIP2017}. But because our face localization relies on HEVC syntax and not on pixel values, our method can still find faces in scrambled images, without being able to recognize them. Hence, we achieve the benefit of enabling simple analytics (such as counting people, estimating their location, etc.) in a privacy-friendly manner, without the need for complex computer vision processing (such as face detection) prior to encoding.

The %remainder of the 
paper is organized as follows. Section~\ref{sec:proposed} presents the proposed face localization method, including feature creation from HEVC syntax.  
%as well as its pixel-domain counterpart, which will be used for benchmarking. 
The scrambling method used to demonstrate privacy-friendly properties of the proposed face localization is briefly described in~\ref{sec:scramble}. Results are presented in Section~\ref{sec:Experiments} followed by conclusions in Section~\ref{sec:conclusion}. 
\vspace{-10pt}
\section{Proposed Method}
\label{sec:proposed}
\begin{figure}[t]
	\centering		
	\begin{subfloat}{}
		\centering
		\centerline{\includegraphics[scale=0.35]{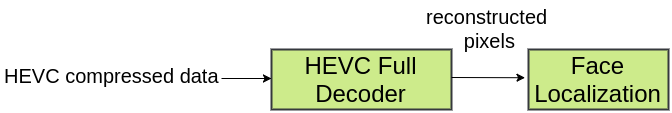}} {(a)} \\
		\label{fig:PD}
	\end{subfloat}
	
	\begin{subfloat}{}	
		\centerline{\includegraphics[scale=0.35]{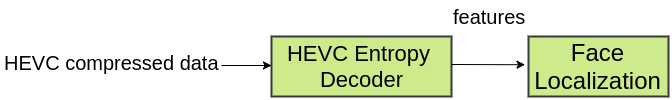}} {(b)} \\
		\label{fig:CD}	
	\end{subfloat}
	\caption{(a) Conventional face localization; (b) proposed face localization.}
	\label{fig:flow_chart}
\end{figure}

Multimedia data is generally only available in  compressed form. Conventional face localization implicitly requires full pixel reconstruction from the compressed data. An example is shown in Fig.~\ref{fig:flow_chart}(a), where the input is an HEVC-compressed image. By contrast, the proposed approach only requires HEVC entropy decoding to reconstruct various syntax elements that will be used as features for face localization. This way,  
%the conventional face localization methods are applied to the output of decoder which fully decodes the encoded bitstream. On the other hand, face localization method based on entropy decoded data achieves significant time savings as it needs only the entropy decoder to run and skips other 
a number of stages in the decoding process can be avoided: inverse quantization, inverse transforms, prediction, and pixel reconstruction. 

%In addition to computational saving in the compression-domain face localization method, we show that if a minor change is made in the encoded bits (e,g. AC coefficient random sign inversion and permutation) the accuracy of the proposed face localization method based on pixel-domain data drops drastically. However, the performance of the face localization method based on entropy decoded data remains unchanged.

In order to perform face localization, we construct a \emph{feature image} from HEVC syntax elements. Specifically, during HEVC entropy decoding, the Intra Prediction Mode (IPM), Prediction Unit Size (PUS) and Bin Number (BN) are reported for each Prediction Unit (PU). We construct a 3-channel feature image based on these parameters by mapping each one to the range 0-255 and copying it into the corresponding location in the feature image. 
 
\begin{figure}[!b]	
	\centering
	\centerline{\includegraphics[scale=0.3]{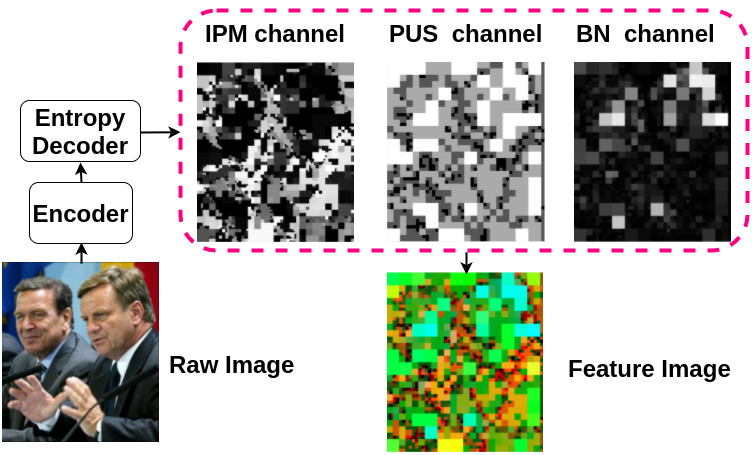}}
	\caption{Creating the feature image.} 
	\label{fig:feature_image}
\end{figure}

IPM values %which are representing DC, planar and directional intra modes 
are integer numbers in the range 0-34~\cite{HEVC}. These are linearly mapped and rounded to integers in 0-255 to create the IPM channel. PUS values can take one of the values $\{4, 8, 16, 32\}$; they are mapped to $\{0, 85, 170, 255\}$, respectively, to create the PUS channel. BN values %are not fixed in a specific range and
vary depending on the number of bits used in a given PU. %The complexity of the encoded PU affects the range of the BN values. 
For the BN channel, the minimum and maximum BN values in the image are found, and then each BN value is linearly mapped and rounded to integers in the range 0-255. %The created 3-channel feature image can be used in any visual analytic task similar to the conventional 3-channel signals. 
An example is shown in Fig.~\ref{fig:feature_image}. Further examples can be found in our earlier work~\cite{MIPR}. 

%Considering the fact that smallest PU is a $4 \times 4$ unit and the reported elements for each channel in feature image are obtained for PUs, we can conclude that the height and width of the feature images should be $\frac{1}{4}$ of the height and width of the input image. But for better visualization and easier comparison to the pixel-domain based face localization we extended the feature image to have same resolution as in the input image.   

%\begin{figure*}[t]	
%	\centering
%	\centerline{\includegraphics[scale=0.35]{./figures/YOLO.png}}
%	\caption{ The modified YOLO9000 network}
%	\label{fig:YOLO}
%\end{figure*} 

We build our face localization upon the state-of-the-art object detector called You Only Look Once (YOLO)~\cite{YOLO}. YOLO is based on a DNN that can find, in a single pass, various objects in the input image along with their bounding boxes. The network is trained to do both object localization and classification using a loss function that includes both bounding box error and class error terms~\cite{YOLO}:   
\begin{multline}
%\begin{split}
%\begin{gathered}
\lambda_{coord} \sum_{i=0}^{S^2} \sum_{j=0}^B \mathbbm{1}_{ij}^{obj} \left[(x_i - \hat{x_i})^2 + (y_i - \hat{y_i})^2 \right] \\
+ \lambda_{coord} \sum_{i=0}^{S^2} \sum_{j=0}^B \mathbbm{1}_{ij}^{obj} \left[(\sqrt{w_i} - \sqrt{\hat{w_i}})^2 + (\sqrt{h_i} - \sqrt{\hat{h_i}})^2\right]  \\
+\sum_{i=0}^{S^2} \sum_{j=0}^B \mathbbm{1}_{ij}^{obj} (C_i - \hat{C_i})^2 +\lambda_{noobj} \sum_{i=0}^{S^2} \sum_{j=0}^B \mathbbm{1}_{ij}^{noobj} (C_i - \hat{C_i})^2 \\ 
+\sum_{i=0}^{S^2} \mathbbm{1}_{i}^{obj} \sum_{c \in classes} ({p_i}(c) - \hat{p_i}(c))^2
%\end{gathered}
%\end{split}
\label{eq:loss}
\end{multline}
where $(x_i,y_i)$ is the center of the ground truth bounding box, $w_i$ and $h_i$ are its width and height, $(\hat{x_i},\hat{y_i})$ is the center of the predicted bounding box whose width and height are $\hat{w_i}$ and $\hat{h_i}$, respectively. $C_i$ and $\hat{C_i}$ are the groud truth and predicted confidence scores corresponding to cell $i$, ${p_i}(c)$ and $\hat{p_i}(c)$ are the ground truth and predicted conditional probabilities for the object class $c$ in cell $i$, $\mathbbm{1}_{ij}^{obj}$ is equal to $1$ if the $j$-th bounding box in cell $i$ is responsible for prediction (i,e. box $j$ has the largest Intersection-over-Union, IoU, among all boxes in cell $i$), and  $\mathbbm{1}_{ij}^{noobj}=1-\mathbbm{1}_{ij}^{obj}$. The scaling factors used are  $\lambda_{coord}=5$ and $\lambda_{noobj}=0.5$. 

The YOLO architecture can be trained to detect different object classes. However, since we are interested in faces only, we used its recent version YOLO9000~\cite{Yolo9000} and modified it to detect one object class - faces. The modified network 
%(Fig.~\ref{fig:YOLO}) 
produces a map of $13\times13$ cells, with each cell returning 5 candidate bounding boxes and a confidence score for each box. The confidence score represents how confident the model is that the corresponding box contains a face. The confidence values can be thresholded to make final predictions: boxes with high enough confidence are predicted to contain faces, and others are ignored. In order to evaluate such a system, a range of thresholds on confidence values is used, and both the prediction accuracy and localization accuracy are taken into account~\cite{FDDB}. The complete evaluation is described in detail in Section~\ref{sec:Experiments}, along with model training.

\section{Privacy Friendliness}
\label{sec:scramble}
%Recent frameworks based on DNN have become very effective in many computer vision tasks. However, DNNs are shown to be not robust to perturbations that may occur as a result of changes in compression, resizing, and cropping corruptions in the images~\cite{improve_DNN}. Since the pixel values are crucial for face localization method based on pixel-domain data, minor variations in intensity of pixels can result in the localization failure. We show that considerable changes in pixel-domain can be made by subtle modifications in the encoded coefficients. To prove this, 
Since our proposed method does not rely on pixel values, it opens up the opportunities for privacy-friendly face localization. To demonstrate this, we adapt the scrambling methods from~\cite{scramble_ICIP} to HEVC. The scrambling schemes in~\cite{scramble_ICIP} were developed to scramble the Region Of Interest (ROI) in H.264/AVC-based video coding. Two basic schemes were proposed: random sign inversion of AC transform coefficients and random permutation of AC coefficients based on the Knuth shuffle~\cite{Knuth_shuffle}.  

We adapt the methods from~\cite{scramble_ICIP} to HEVC and apply them across the entire image. Since the transform coefficients are computed in each Transform Unit (TU), we apply random sign inversion and random permutation within each TU. These changes can be undone by an authorized decoder that knows the (pseudo)random sequences involved in sign inversion and coefficient permutation. An unauthorized decoder will only be able to reconstruct scrambled images. 

The above changes have a significant effect on the final reconstructed images, rendering conventional face localization (and presumably face recognition) useless. However, they have only a minor effect on our feature images, hence our face localization is largely unaffected by such scrambling. Specifically, the IPM and PUS channel remain unaffected. Random permutation and sign changes do increase BN values over the whole image. But because BN channel is produced by normalizing BN values using the minimum and maximum BN values in the image, the net effect on the BN channel is minor. We measured the mean of absolute intensity difference in the BN channel between scrambled and non-scrambled images, and found that the difference is only 0.1, averaged over all the training images. 

\begin{figure}[t]	
	\centering
	\centerline{\includegraphics[scale=0.28]{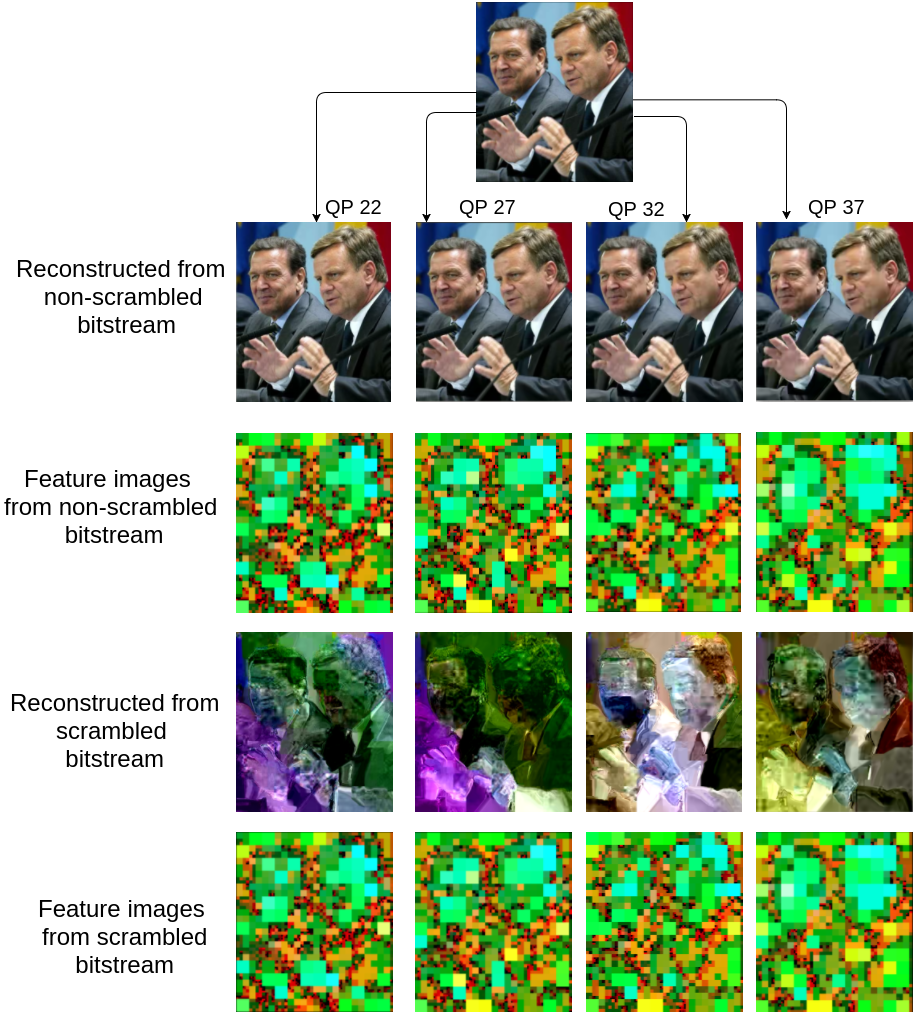}}
	\caption{An example of feature images and fully reconstructed images for the input encoded with no-scramling case and with scrambling  for 4 different QP values.} 
	\label{fig:QP_effect}
\end{figure}

An example of the effects of scrambling is shown in Fig.~\ref{fig:QP_effect} for four quantization parameter (QP) values. One can see that scrambling has a major effect on the final reconstructed images, but not on our feature images. Therefore, using our approach, one would still be able to detect and localize faces in the scrambled bitstreams, but would not be able to reveal their identity.

\vspace{-13pt}
\section{Experimental Results}
\label{sec:Experiments}
\subsection{Experimental Setting}
\label{subsec:EXP_setting}
Face Detection Dataset and Benchmark (FDDB)~\cite{FDDB} is used for evaluating the performance of the proposed face localization method. FDDB includes 2845 images with 5171 annotated faces. FDDB comes with a standard evaluation method that allows comparison among various face localization methods. The evaluation is based on the Intersection-over-Union (IoU) with the ground truth; if IoU is larger than $0.5$, the detection is considered a True Positive (TP)~\cite{FDDB}. 
 
HEVC reference software HM16.5~\cite{HM} is used for intra coding the images using the configurations in~\cite{hevc_ctc}. The QP values used in the evaluation are $\{22, 27, 32, 37\}$, which covers the range typically used in practice. 
 
Our CNN model is based on the Darknet framework~\cite{darknet}. The training data were the feature images extracted from the (non-scrambled) HEVC bitstream obtained with QP $=32$. Stochastic gradient descent with learning rate $10^{-3}$, momentum of $0.9$, and weight decay of $5 \times 10^{-4}$ is used for training. The training batch size was set to $64$, and the training was terminated after 10k epochs. %NVIDIA P100 Pascal GPUs are used for training and testing. 
 Training was initialized with YOLO900 model weights obtained on ImageNet~\cite{imagenet}. For testing, non-maximum suppression~\cite{NMS_PAMI_2010} on the outputs with the threshold of $0.4$ was employed. 
 
\begin{figure}[t]	
	\centering
	\centerline{\includegraphics[width=8cm]{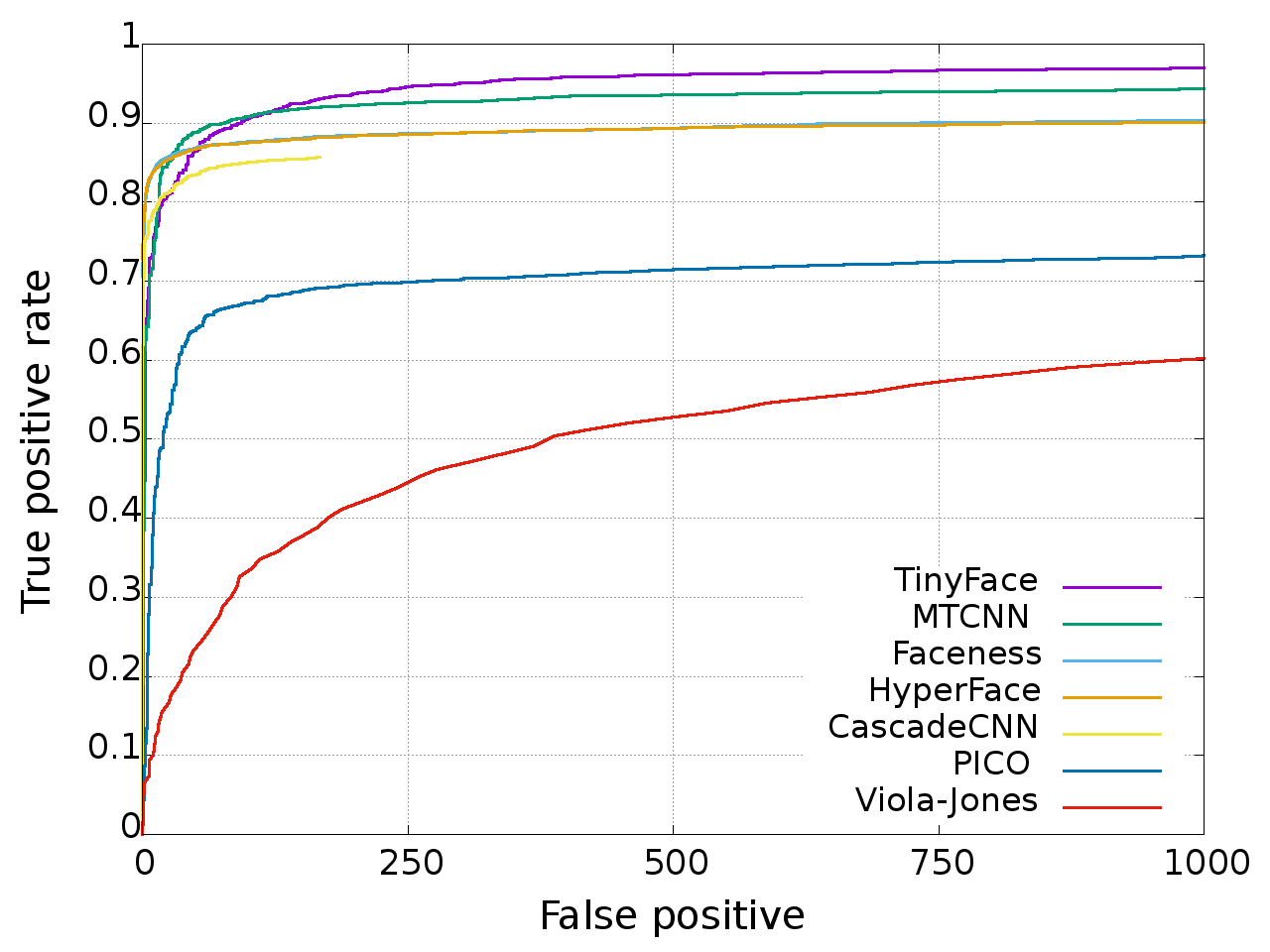}}
	\caption{Several notable face localization models on FDDB, including the chosen benchmark.} 
	\label{fig:PD_ROC_1}
\end{figure}

Fig.~\ref{fig:PD_ROC_1} shows the performance of several notable face localization methods on FDDB, including TinyFace~\cite{tiny}, MTCNN~\cite{MTCNN}, Faceness~\cite{facial_parts}, Hyperface~\cite{hyperface}, CascadeCNN~\cite{cascade}, PICO~\cite{PICO} and Viola-Jones~\cite{Viola-jones}. We have chosen TinyFace as the benchmark to compare against, since it represents the current state-of-the-art.
%\subsection{Results of Face Localization for Pixel-domain Data}
%\label{subsec:EXP_PD}
%To compare the performance of proposed method with the state-of-art methods, the CNN model (Fig.~\ref{fig:YOLO}) for face localization using pixel-domain data is trained with the images in the dataset without applying compression. The performance of the proposed face localization method in Section \ref{subsec:PD_detector} is evaluated through 10-fold cross validation method. 

%The comparison of the results with the state-of-art face localization methods is shown in Fig~\ref{fig:PD_ROC_1}. As it can be seen in the figure, the proposed method based on the pixel-domain data achieves higher TP rate for the given False Positive (FP) numbers compared to many of the state-of-art face localization methods.

%Main advantage of the proposed face localization method is the simplicity of the face localization technique. There is no preprocessing such as partitioning the input into patches. We do not use ensemble averaging using multiple models in the proposed method and no complex post processing is applied to the output of the CNN model. As a result, the proposed method can be easily used in real-time applications.   

%The same trained models are used for testing the reconstructed sequences encoded with QP values $\in \{22, 27, 32, 37\}$ to check the robustness of the proposed method to encoding with different QP values. The comparison of the results with different QP values is given in Fig~\ref{fig:PD_ROC_2}. As it is illustrated in the figure, the performance of the proposed method based on the pixel-domain data does not change for the test sequences encoded with QP $\in \{22, 27, 32\}$. However, there is a slight drop in the TP rate in tests with QP=37 due to the higher distortion level in larger QP values.    
%\begin{figure}[!t]	
%	\centering
%	\centerline{\includegraphics[scale=0.20]{./figures/QP_effect_PD.png}}
%	\caption{The performance of the proposed pixel-domain based method tested with the test sequences reconstructed based on the encoded bitstreams obtained with QP $\in \{22,27,32,37\}$ on the FDDB dataset.} 
%	\label{fig:PD_ROC_2}
%\end{figure}
\vspace{-13pt}
\subsection{Face localization results}
\label{subsec:EXP_CD}

\begin{figure}[t]	
	\centering
	\centerline{\includegraphics[width=8.0cm]{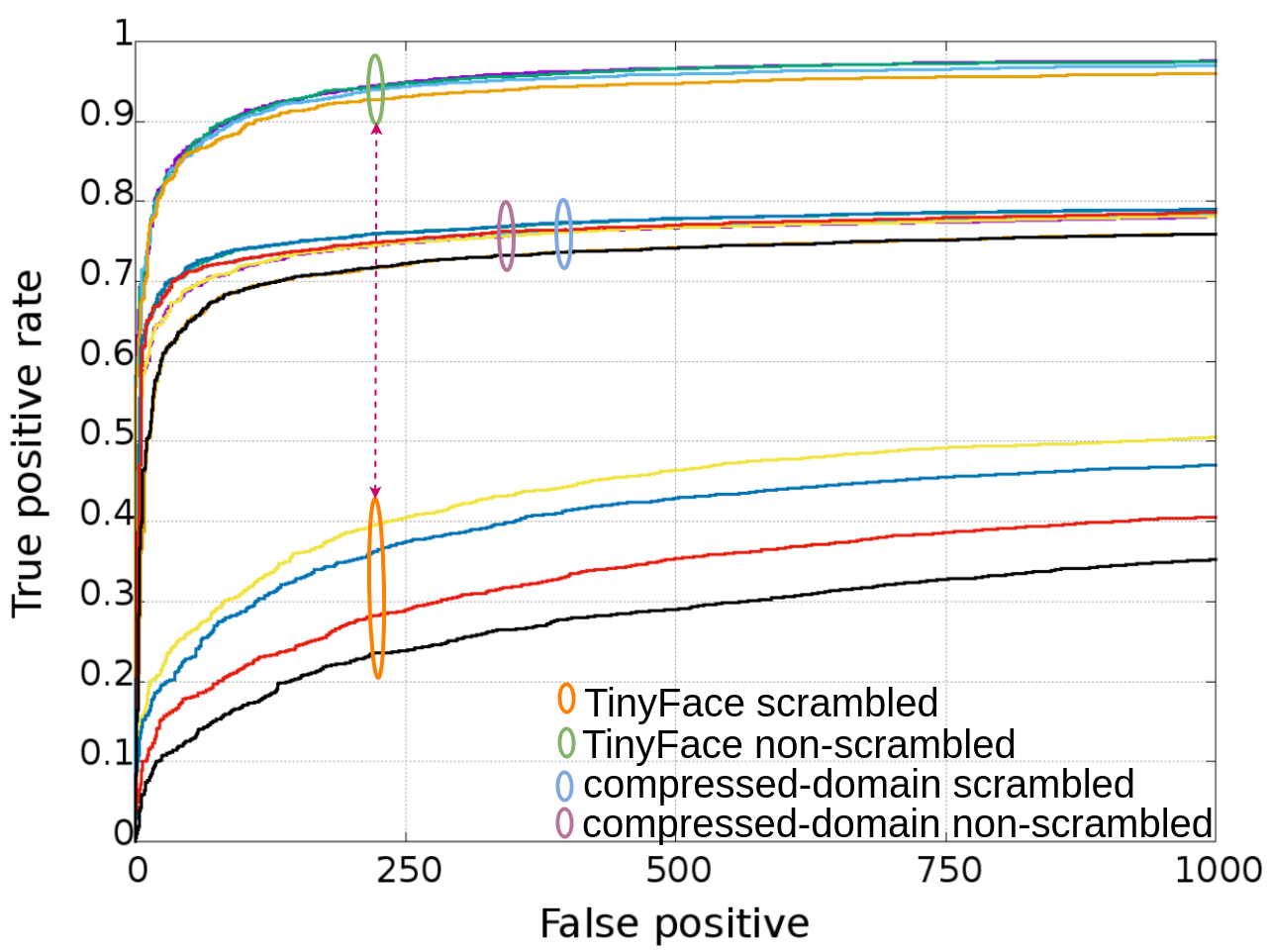}}
	\caption{At FP=1000, TinyFace~\cite{tiny} has over 97\% TP rate on raw images as well as images decoded from non-scrambled bitstreams. But on images decoded from scrambled bitstreams, its TP rate drops to 50\% or less. Meanwhile, our method has consistent performance on both scrambled and non-scrambled images. The different colors correspond to different QP values. } 
	\label{fig:CD_ROC_1}
\end{figure}

The test data consists of FDDB images encoded in the HEVC intra mode using QP $\in \{22,27,32,37\}$, as mentioned before. We used both scrambled and non-scrambled bitstreams to investigate the effect of scrambling on face localization. The input to our model were feature images generated as described in Section~\ref{sec:proposed}. The input to TinyFace were the fully-decoded images. The FDDB accuracy results are shown in Fig.~\ref{fig:CD_ROC_1}.  

%A training process similar to what is discussed in Section \ref{subsec:EXP_PD} is performed for the method based on the compression-domain data. However, the images that are used for training the designed CNN are the feature images obtained using the entropy decoded bitstreams. As mentioned in Section \ref{subsec:CD_detector}, the feature images change when the encoding QP value changes. As a result, for each QP value a separate network is trained and tested with the feature images corresponding to the same QP. The ROC curves for each experiment is shown in Fig~\ref{fig:CD_ROC_1}. Similar to the face localization method based on the pixel-domain data, the obtained curves are very similar for test sequences encoded with QP $\in \{22, 27, 32\}$ and for QP=37 the TP rate for the given FP is slightly lower compared to the remaining QP values. For larger QP, the rate distortion optimization (RDO) process tries to avoid choosing smaller partitions due to the required additional bits for signaling smaller partitions. Hence, the finer partitions start to diminish in the feature images  as the QP increases and the smaller faces in the test images become more challenging to be localized.

\begin{figure}[b!]	
	\centering
	\centerline{\includegraphics[width=8.1cm]{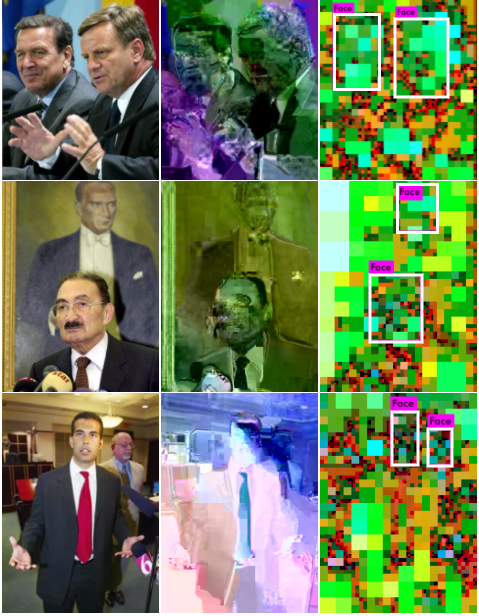}}
	\caption{TinyFace~\cite{tiny} cannot find  faces in the scrambled images, but our model finds faces in feature images extracted from scrambled bitstreams.}
	\label{fig:visual_results}
\end{figure} 

As seen in Fig.~\ref{fig:CD_ROC_1}, the benchmark model achieves over 97\% TP rate with 1,000 False Positives (FP) on non-scrambled images. The effect of HEVC compression on this model is minor in the range of QP values we used. But on scrambled images, its TP rate drops to 50\% or less. This shows the significant effect that scrambling has on face localization, as we could have expected from Fig.~\ref{fig:QP_effect}. 

Meanwhile, our model achieves roughly the same performance on both scrambled and non-scrambled images, as discussed in Section~\ref{sec:scramble}. For our model, QP has a larger influence than scrambling, because feature images change with QP (Fig.~\ref{fig:QP_effect}). While we could have trained an ensemble of models, each on a different QP, for possibly better performance, we opted to use a single model trained on QP $=32$ in order to examine its robustness to QP variation. Indeed, even when tested on data produced by different QPs, our model still outperforms the benchmark significantly on privacy-friendly scrambled images. It is worth noting that neither the proposed method nor the benchmark were trained on scrambled images. 

%\textcolor{blue}{ We expect that if they were trained on scrambled images, their performance would likely improve, so long as the test images were scrambled in the same way as training images. Hence, for this type of training, advance knowledge of the scrambling procedure would be required. In the experiments shown in Fig.~\ref{fig:CD_ROC_1} the proposed method and the benchmark are trained using the non-scrambled data. It is expected that if TinyFace is trained using the reconstructed scrambled bitstreams, its performance would improve when tested with scrambled sequences. However, the TinyFace performance is still highly dependent on the reconstructed pixel values. This means there will be a significant drop in the performance of TinyFace if an additional minor change in the coefficients is applied.} 
%\textcolor{red}{Here we need an impressive figure showing various FDDB images (original, scrambled with output from benchmark, scrambled with output from our model). This figure could stretch over two columns if needed. It also needs to be explained in the text. }

Fig.~\ref{fig:visual_results} shows a few examples of face localization. In each row, the first image is the original, the second image is decoded from the scrambled bitstream and the last image is the feature image extracted from the scrambled bitstream. TinyFace~\cite{tiny} cannot find  faces in scrambled images, but our model finds faces in feature images extracted from scrambled bitstreams. 
\vspace{-15pt}
\section{Conclusion}
\label{sec:conclusion}
%\textcolor{red}{Still to be revised.} In this paper, face localization methods based on pixel-domain data and compression domain data are proposed. The performance of the proposed simple face localization method based on pixel-domain data is comparable to the state-of-art face localization methods based on CNN. The main contribution of the paper is the face localization approach based on the compression-domain data which achieves reasonably hight detection rates requiring less than half of the decoding time compared to the pixel-domain based method. 

%Although the method based on pixel-domain data achieves higher accuracy, its performance drops significantly when a simple scrambling is applied. On the other hands, the method based on the compression-domain data is proved to be robust to the applied scrambling. This feature will be studied in future works to find methods for privacy-preserving face localization methods.  

In this paper we presented a method for finding faces in HEVC-coded images. Our approach takes advantage of HEVC syntax, rather than the actual pixel values, which allows it to find faces even in scrambled images. This opens up the possibilities for privacy-friendly visual analytics, such as counting people without revealing their identity. Unlike existing approaches, our methodology does not require running complex computer vision engines prior to encoding. 

The proposed method runs on HEVC intra-coded bitstreams, primarily because still images are the common setting for evaluating face detectors/localizers~\cite{FDDB}. But the methodology lends itself to extension to video as well. In the case of HEVC-coded video, faces could be detected and localized in I-frames, then tracked through the inter-coded frames using motion vectors, for example using~\cite{kb_tip_2013}. 

\bibliographystyle{IEEEbib}
\bibliography{refs}

\end{document}